\documentclass[3p,twocolumn]{elsarticle}

\usepackage{amssymb}
\usepackage{tabularx}
\usepackage{threeparttable}
\usepackage{multirow}
\usepackage{amsfonts} 
\usepackage{algorithm}
\usepackage{algorithmic}
\usepackage{bbding}
\usepackage{setspace}
\usepackage{makecell}
\usepackage{booktabs}
\usepackage{adjustbox}
\usepackage{amsmath}

\journal{arxiv}

\begin{document}

\begin{frontmatter}

\title{PGDA-KGQA: A Prompt-Guided Generative Framework with Multiple Data Augmentation Strategies for Knowledge Graph Question Answering}

\author[info1]{Xiujun Zhou}
\ead{2840646724@qq.com}

\author[info1]{Pingjian Zhang}
\ead{pjzhang@scut.edu.cn}

\author[info1]{Deyou Tang\corref{cor1}}
\ead{dytang@scut.edu.cn}
\cortext[cor1]{Corresponding author.}

\affiliation[info1]{organization={School of Software Engineering},
            addressline={South China University of Technology}, 
            city={Guangzhou},
            postcode={510641}, 
            country={China}}

\begin{abstract}
Knowledge Graph Question Answering (KGQA) is a crucial task in natural language processing that requires reasoning over knowledge graphs (KGs) to answer natural language questions. Recent methods utilizing large language models (LLMs) have shown remarkable semantic parsing capabilities but are limited by the scarcity of diverse annotated data and multi-hop reasoning samples. Traditional data augmentation approaches are focus mainly on single-hop questions and prone to semantic distortion, while LLM-based methods primarily address semantic distortion but usually neglect multi-hop reasoning, thus limiting data diversity. The scarcity of multi-hop samples further weakens models' generalization. To address these issues, we propose PGDA-KGQA, a \textbf{p}rompt-guided \textbf{g}enerative framework with multiple \textbf{d}ata \textbf{a}ugmentation strategies for KGQA. At its core, PGDA-KGQA employs a unified prompt-design paradigm: by crafting meticulously engineered prompts that integrate the provided textual content, it leverages LLMs to generate large-scale (question, logical form) pairs for model training. Specifically, PGDA-KGQA enriches its training set by: (1) generating single-hop pseudo questions to improve the alignment of question semantics with KG relations; (2) applying semantic-preserving question rewriting to improve robustness against linguistic variations; (3) employing answer-guided reverse path exploration to create realistic multi-hop questions. By adopting an augment-generate-retrieve semantic parsing pipeline, PGDA-KGQA utilizes the augmented data to enhance the accuracy of logical form generation and thus improve answer retrieval performance. Experiments demonstrate that outperforms state-of-the-art methods on standard KGQA datasets, achieving improvements on WebQSP by 2.8\%, 1.2\%, and 3.1\% and on ComplexWebQuestions by 1.8\%, 1.1\%, and 2.4\% in F1, Hits@1, and Accuracy, respectively.
\end{abstract}


\end{frontmatter}

\section{Introduction}

KGQA is a fundamental task in natural language processing that enables machines to answer natural language questions using structured knowledge from KGs like DBpedia \cite{auer2007dbpedia}, Freebase \cite{bollacker2008freebase} and Wikidata \cite{vrandevcic2014wikidata}. These KGs are typically represented as factual triples in the form of (head entity, relation, tail entity). Existing KGQA approaches primarily addressed two key issues: knowledge retrieval \cite{yao2007knowledge} and semantic parsing \cite{berant2013semantic}. Knowledge retrieval focuses on identifying  relevant entities, relations or triples from the KG to narrow down the candidate space. In contrast, semantic parsing converts unstructured questions into structured logical forms \cite{gu2021beyond}, which can be translated into executable queries like SPARQL \cite{perez2009semantics} to obtain accurate answers and interpretable reasoning paths.

Advances in LLMs have greatly enhanced natural language understanding and generation. However, ensuring factual reliability remains a critical issue. Although LLMs acquire factual knowledge from large-scale corpora, their ability to accurately recall this knowledge is often limited, leading to factual inconsistencies or hallucinations \cite{ji2023survey, chen2024multi} in downstream tasks such as question answering. To handle this problem, retrieval-augmented generation (RAG) \cite{lewis2020retrieval} effectively enhances LLMs performance by integrating external knowledge, improving factual accuracy, and reducing hallucinations \cite{shuster2021retrieval, tonmoy2024comprehensive}. Among knowledge sources, KGs excel due to their structured fact representation as compact triples, enabling symbolic reasoning with greater interpretability \cite{zhang2021neural}. Recent research has focused more on integrating KGs with LLMs, leveraging their strengths and leading to notable KGQA improvements with more accurate and interpretable answers \cite{jiang2023structgpt, sun2024thinkongraph}.

Semantic Parsing (SP) based approaches \cite{hu2022logical, ye2022rng, zhang2023fc} aim to translate natural language questions into structured queries (e.g., SPARQL) that can be executed over KGs for accurate answer retrieval. Nevertheless, their effectiveness is largely constrained by the availability of high-quality SPARQL annotations. While state-of-the-art SP-based methods \cite{luo2024chatkbqa, feng2025rgr} demonstrate excellent semantic parsing capabilities in KGQA, accurately generating logical forms remains challenging when encountering unseen entities or relations, diverse linguistic expressions, and complex multi-hop questions. 

Traditional approaches rely on domain experts to design rule-based templates or annotate large datasets. Recently, LLMs have been used to convert triples and reasoning paths into natural language, reducing semantic gaps from conventional augmentation methods like synonym substitution and paraphrasing. However, these methods tend to focus on single-hop questions and primarily enhance semantic understanding, often neglecting the generation of diverse, high-quality SPARQL-annotated data for multi-hop reasoning \cite{wu2023retrieve}. 

To resolve this difficulty, we propose PGDA-KGQA, a prompt-guided generative framework designed to enhance SP-based KGQA by leveraging LLMs to generate diverse, high-quality SPARQL-annotated data. Built upon an augment-generate-retrieve semantic parsing pipeline, PGDA-KGQA applies three data augmentation strategies to enhance both single-hop and multi-hop reasoning: 
(1) Single-hop Pseudo Question Generation (SPQG): Retrieved KG triples and prompt templates guide LLMs to generate diverse question templates for each relation. By replacing entity placeholders, we produce pseudo questions with logical forms, improving the model's ability to map natural language to KG relations.
(2) Semantic Preserving Question Rewriting (SPQR): We employ LLMs to rephrase questions into multiple variations while preserving their original meaning, improving the model's understanding of diverse expressions. 
(3) Answer-guided Reverse Path Exploration (ARPE): Unlike prior methods that generate reasoning paths and answer nodes via random walks \cite{jiang2023reasoninglm}, ARPE first designates answer nodes and then explores reasoning paths in reverse, based on different reasoning patterns, generating a large number of semantically meaningful questions.

Our contributions are as follows: (1) We propose PGDA-KGQA, a prompt-guided augment-generate-retrieve framework that leverages meticulously designed prompts to guide LLMs in generating large-scale SPARQL-annotated (question, logical form) pairs, offering a practical solution for semantic parsing and facilitating multi-hop question answering. (2) We introduce three LLM-based data augmentation strategies: single-hop pseudo question generation, semantic-preserving question rewriting, and answer-guided reverse path exploration. These strategies independently enhance model generalization by improving alignment with KG relations, robustness to linguistic variations, and the ability to handle multi-hop reasoning. (3) We carry out experiments on WebQSP \cite{yih2016value} and ComplexWebQuestions (CWQ) \cite{talmor2018web}, the results demonstrate the effectiveness of our methods. Integrated into an LLM-based augment-generate-retrieve KGQA framework, each strategy contributes to improved KGQA performance, achieving new state-of-the-art results.

\section{Related works}
\label{sec2}

KGQA involves answering questions by retrieving factual knowledge from a knowledge graph. In general, existing KGQA methods fall into two main categories: information retrieval (IR)-based and semantic parsing (SP)-based approaches. To bridge the gap between structured KG representations and natural language, KG-to-text techniques have been explored, which provide a foundation for generating diverse training data.

\textbf{IR-based Methods} typically retrieve relevant factual triples from a KG to construct subgraphs, which are then used within RAG frameworks to derive answers. Prior research \cite{sun2019pullnet, das2022knowledge, zhang2022subgraph, baek2023knowledge} has explored subgraph retrieval for answer generation. Recently, \citet{sun2024thinkongraph} treats subgraph retrieval as a reasoning task, employing an LLM to generate reasoning steps.

\textbf{SP-based Methods} aim to transform a question into an executable logical query that can be executed directly on a KG to retrieve answers. Some studies \cite{lan2020query, chen2023outlining} use techniques for step-by-step query graph generation and search in semantic parsing. Alternatively, other SP-based approaches \cite{das2021case, ye2022rng, shu2022tiara, luo2024chatkbqa, feng2025rgr} focus on generating logical forms directly from questions. Despite the advanced generative capabilities of LLMs, these methods still leave room for improvement, particularly in terms of accuracy.

\textbf{KG-to-text} techniques convert triples and reasoning paths from KGs into natural language text. For example, \citet{atif2023beamqa} samples triples from the input graph and connects the entity and relation of a triple to form a question. \citet{huang2024less} generates single-hop questions from triples using LLMs to train autoregressive models for relation path prediction. \citet{jiang2023reasoninglm} improves subgraph understanding by performing random walks on the KG, designating the final entity as the answer, and using rule-based templates and LLMs to generate questions. However, there is still a lack of methods for generating SPARQL-annotated data for SP-based methods.

\section{Preliminaries}

In this section, we present two basic concepts of our work: knowledge graph, logical form, followed by the problem statement for KGQA tasks.

\textbf{Knowledge Graph}. A KG is typically structured using the Resource Description Framework (RDF) and consists of triples $(s, r, o)$, where $s$ is an entity, $o$ is either an entity or a literal, $r$ denotes the relation between them. Formally, a KG is defined as $KG = \{(s,r,o) \mid s \in E, r \in R, o \in E \cup L\}$. Each entity $e$ in the entity set $E$ is represented by a unique ID (e.g. $e$.id = ``m.01xpjyz'') and a queryable label ($e$.label = ``Airport''). Relations in the relation set $R$ follows a structured naming convention, such as ``people.marriage.spouse''. A literal $l \in L$ typically represents numbers (e.g., ``1.0'') or dates (``1999-12-31'').

\textbf{Logical Form}. Logical forms provide a structured representation of natural language questions, offering a more concise alternative to SPARQL, the standard query language for RDF data. Among various logical forms, S-expression \cite{gu2021beyond} is widely used. In this work, we adopt it as the output logical form for our model. There is a formal mapping between S-expressions and SPARQL queries, allowing direct translation. In S-expressions, query logic is represented using function composition: [JOIN $r$ $o$] retrieves the head entity $s$, while [JOIN [ R $r$] $s$] retrieves the tail entity $o$. In this paper, ``logical form'' refers specifically to S-expressions in our methodology.

\textbf{Problem Statement}. For the KGQA task, given a natural language question $Q$ and a knowledge graph $KG$, we first transform $Q$ into a logical form $F$=$SP(Q)$ using the semantic parsing function $SP(\cdot)$. This logical form is then converted into an equivalent SPARQL query $sparql$=$Convert(F)$ through the conversion function $Convert(\cdot)$. Finally, executing $sparql$ on $KG$ produces the final answer set $A$=$Execute(sparql \mid KG)$, where $Execute(\cdot)$ refers to the query execution function.

\section{Methodology}

In this section, we first provide an overview of the PGDA-KGQA framework. Then, we describe three LLM-based data augmentation strategies, fine-tuning open-source LLMs, generating logical forms with fine-tuned LLMs, unsupervised entity and relation retrieval, and explainable query execution.

\subsection{Overview of PGDA-KGQA}

\begin{figure*}[t]
    \centering
    \includegraphics[width=\textwidth]{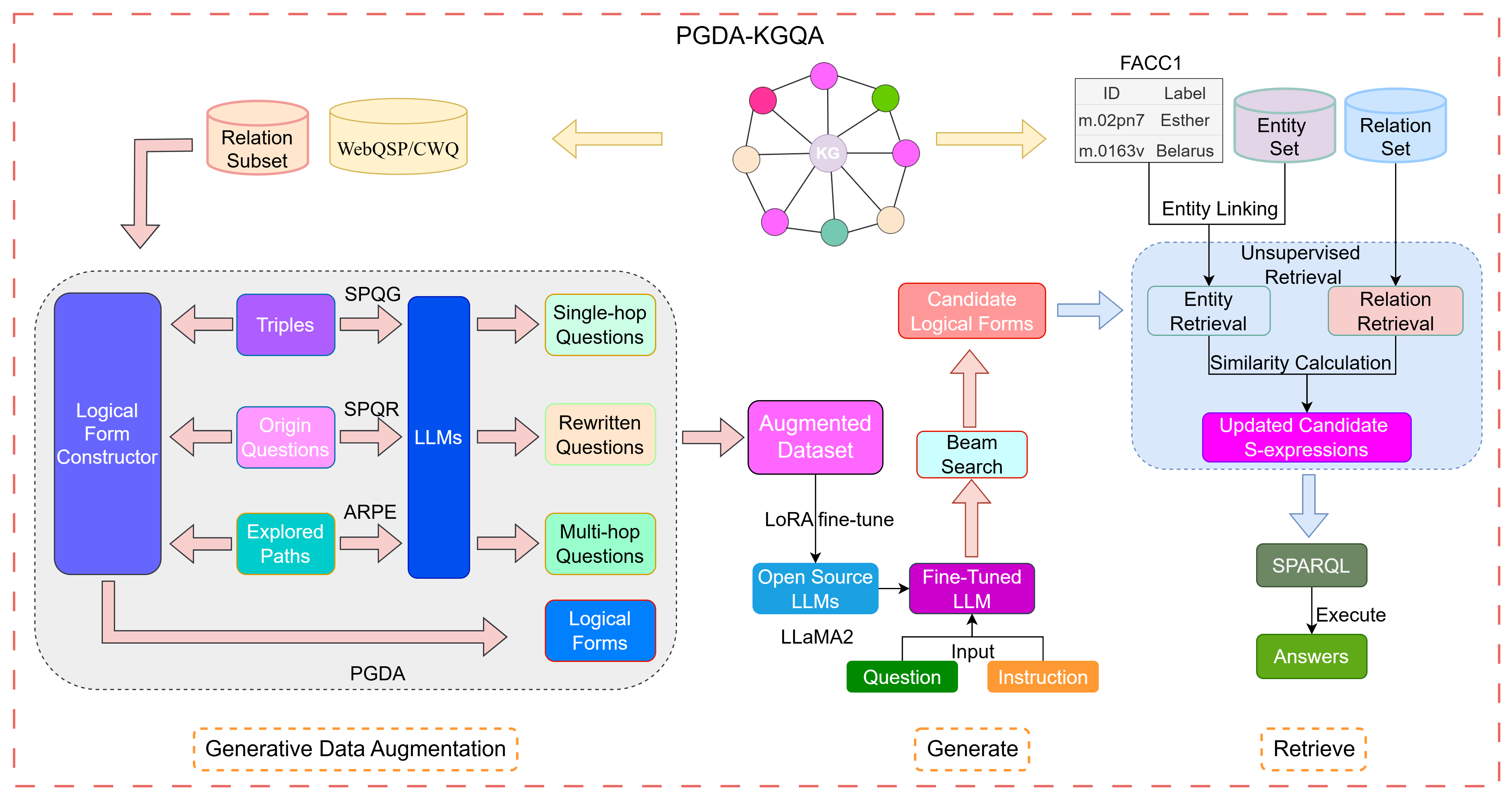}
    \caption{\label{fig:PGDA}The overview of the PGDA-KGQA framework for the augment-generate-retrieve KGQA method. The framework consists of three stages: generative data augmentation, fine-tuning an LLM to generate logical forms, updating the logical forms through unsupervised retrieval, and finally executing SPARQL queries to produce the answers.}
\end{figure*}

As shown in Figure \ref{fig:PGDA}, PGDA-KGQA is a prompt-guided generative framework with data augmentation strategies for KGQA based on fine-tuned LLMs. Given a dataset and a KG, the framework employs three distinct strategies, each driven by a specifically crafted prompt, to generate questions using LLMs and constructs corresponding logical forms. It then fine-tunes an LLM on an augmented dataset comprising <question, logical form> pairs. The fine-tuned LLM generates high-quality logical forms via beam search, followed by unsupervised entity and relation retrieval to refine them. Finally, the ranked logical forms are iteratively converted into SPARQL queries and executed on the KG to obtain the final answers.

\subsection{Data augmentation}

\subsubsection{Single-hop pseudo question generation}

To address the limited coverage in KGQA datasets, we propose SPQG, an LLM-based strategy for generating single-hop pseudo questions. First, we retrieve 8,321 valid relations from the KG and extract 33,262 corresponding entities from the related triples. LLMs are then employed to generate a large number of pseudo questions from these triples, expanding the entity and relation coverage of the training set. This generated data allows the model to learn accurate semantic mappings between natural language questions and KG elements, thus improving KGQA performance.

\renewcommand{\algorithmicrequire}{\textbf{Input:}}
\renewcommand{\algorithmicensure}{\textbf{Output:}}
\begin{algorithm}[!t]
    \small
    \caption{SPQG Algorithm}
    \label{alg:generate_pseudo}
    \begin{algorithmic}[1]
    \REQUIRE A relation subset of Freebase \texttt{$R$}, number of pseudo question templates \texttt{$k$}, prompt \texttt{$p$}
    \ENSURE Dataset Generated by SPQG Strategy \texttt{$D$}

    \STATE \texttt{$D$} $\leftarrow \emptyset$;  \texttt{// Initialize a list}
    \STATE \texttt{$rels$} $\leftarrow \emptyset$;  \texttt{// Initialize a list}
    
    \STATE \textbf{foreach} \texttt{$r$} $\in$ \texttt{$R$} \textbf{do}
        \STATE \quad \texttt{$T$} $\leftarrow$ \texttt{get\_rels\_with\_triples}(\texttt{$r$});
        \STATE \quad \textbf{if} \texttt{len($T$)} > 0 \textbf{then}
            \STATE \quad \quad \texttt{$rels$}.\texttt{append}(\texttt{$r$});
    
    \STATE \texttt{$rel2subjects$} $\leftarrow$ \{\}; \texttt{// Initialize a dictionary}
    \STATE \texttt{$rel2triples$} $\leftarrow$ \{\}; \texttt{// Initialize a dictionary}
    \STATE \textbf{foreach} \texttt{$r$} $\in$ \texttt{$rels$} \textbf{do}
        \STATE \quad \texttt{$rel2subjects$}[\texttt{$r$}] $\leftarrow$ \texttt{get\_rel2subjects}(\texttt{$r$}, \texttt{$k$});   
    
        \STATE \quad \texttt{$rel2triples$}[\texttt{$r$}] $\leftarrow$ \texttt{get\_rel2triple}(\texttt{$r$});

    \STATE \textbf{foreach} \texttt{$r$} $\in$ \texttt{$rels$} \textbf{do}
        \STATE  \quad \texttt{$templates$} $\leftarrow$ \texttt{LLM}(\texttt{$r$}, \texttt{$rel2triples$}[\texttt{$r$}], \texttt{$p$}, \texttt{$k$});

        \STATE \quad \texttt{QA\_pairs} $\leftarrow$ \texttt{build}(\texttt{$templates$}, \texttt{$rel2subjects$}[\texttt{$r$}]);
        \STATE \quad \texttt{$D$}.\texttt{append}(\texttt{QA\_pairs});
    
    \STATE \textbf{return} \texttt{$D$};
\end{algorithmic}
\end{algorithm}

\begin{figure}[ht]
	\centering	
	\includegraphics[width=\columnwidth]{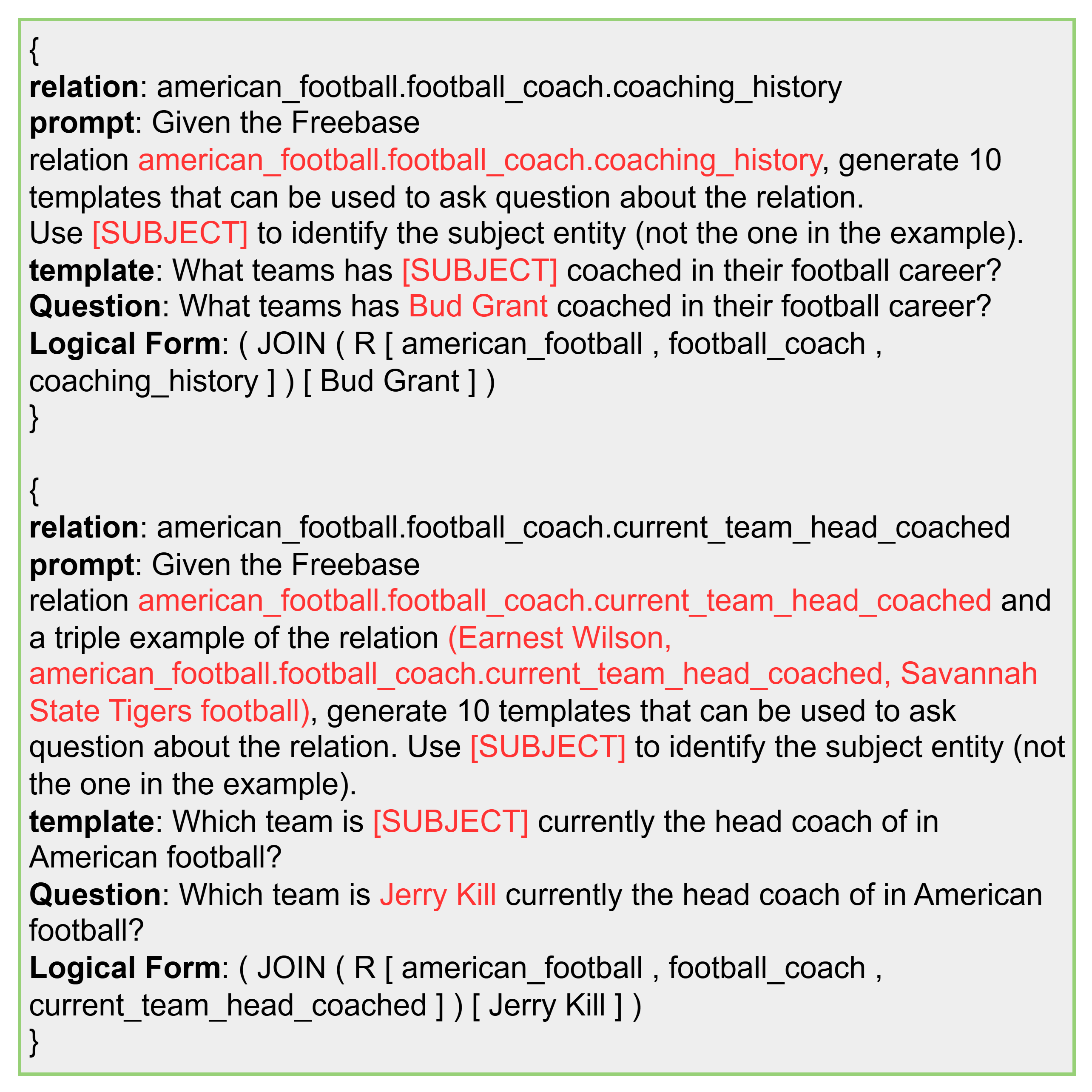} 
	\caption{\label{fig:pseudo_example}Two examples of single-hop question generation by SPQG using different prompts.}
\end{figure}

Algorithm~\ref{alg:generate_pseudo} presents the SPQG algorithm in detail. The inputs include a subset $R$ of Freebase relations\footnote{https://developers.google.com/freebase} (26,282 relations), a parameter $k$ specifying  the number of question templates per relation, and a prompt template $p$ for the LLM. For each $r$ in $R$, we retrieve a relevant triple using \verb|get_rels_with_triples| and retain only relations with at least one valid triple to form $rels$. Then, for each relation in $rels$, we sample $k$ head entity IDs and labels via \verb|get_rel2subjects| while extracting head and tail entity labels using \verb|get_rel2triple|. Using this information, we design the prompt $p$ and use LLMs to generate $k$ question templates, with \verb|[SUBJECT]| representing the subject entity. Finally, \verb|build| substitutes the placeholder with actual head entities from $rel2subjects$ and constructs logical forms for the pseudo questions, creating a dataset of \verb|(question, logical form)| pairs. Figure~\ref{fig:pseudo_example} illustrates two examples of single-hop question generation by SPQG using different prompts.

\subsubsection{Semantic preserving question rewriting}

Existing datasets like WebQSP and CWQ provide only one question per user intent, lacking richness of natural language. To overcome this limitation, we propose the SPQR strategy, which uses LLMs to generate multiple linguistic rephrasings for each original question, improving the model's ability to generalize across diverse question formulations. 

\begin{figure}[t]
	\centering	
	\includegraphics[width=\columnwidth]{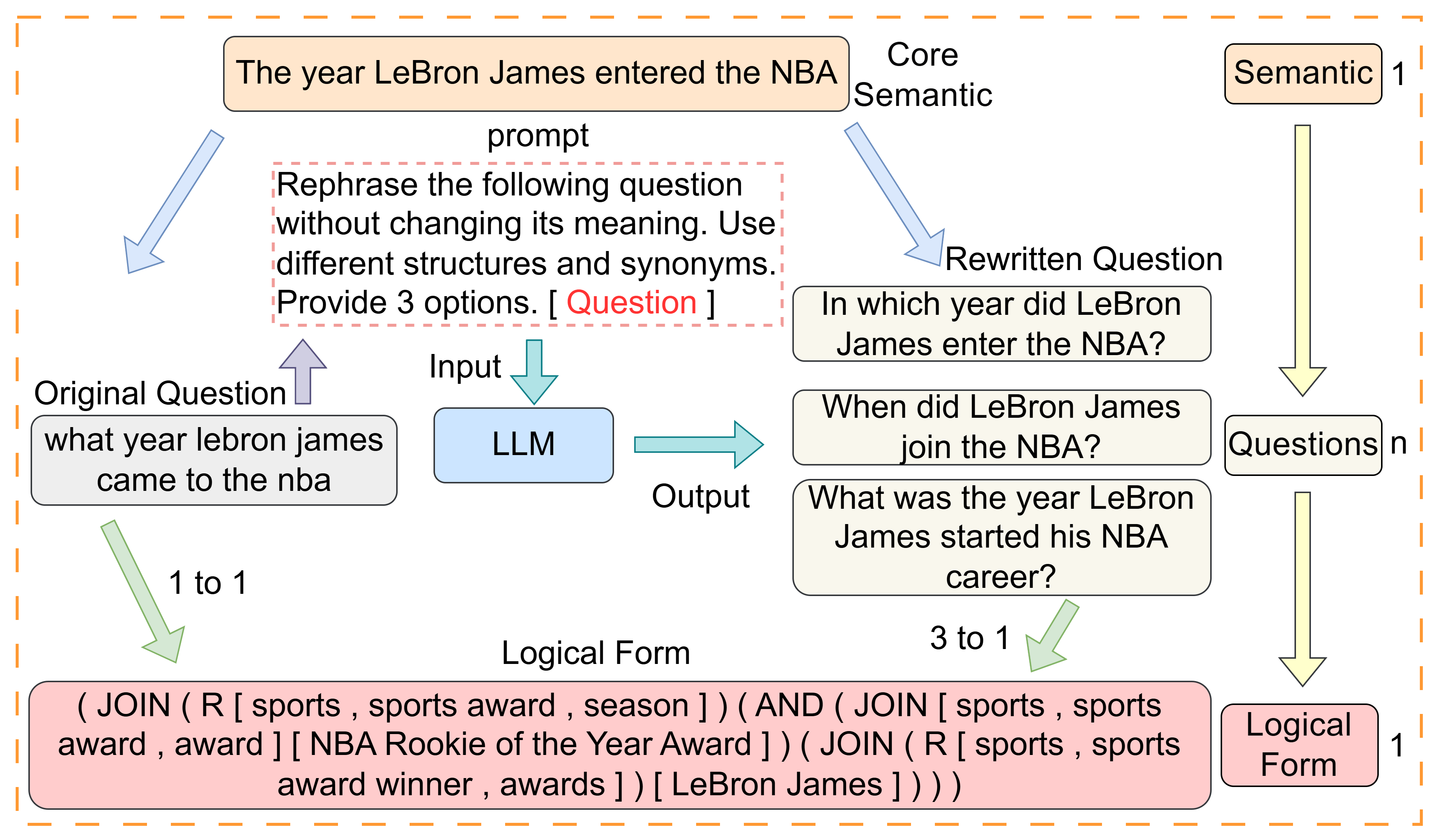} 
	\caption{\label{fig:rewritten_question}An example of rephrasing questions while preserving their meaning.}
\end{figure}

As shown in Figure \ref{fig:rewritten_question}, SPQR uses a prompt to guide LLMs to generate $rw$ distinct rephrasings of each question with different structures and synonyms, while preserving the original semantics. This establishes mappings between multiple rephrasings of the same core meaning and a single logical form, effectively mitigating model overfitting to specific expression forms.

\subsubsection{Answer-guided reverse path exploration}

Existing methods typically use random walks that start at an entity node and traverse a fixed number of hops to identify the answer node, constituting a forward exploration process. In contrast, our ARPE strategy fixes the answer node and reasoning path pattern first, then performs reverse exploration under these constraints. This method guarantees meaningful generated questions by predetermining the answer and allows us to control the reasoning path so that the questions reflect real-world distribution patterns. Consequently, ARPE produces questions that more accurately represent practical scenarios while significantly increasing training samples for rare reasoning patterns.

\renewcommand{\algorithmicrequire}{\textbf{Input:}}
\renewcommand{\algorithmicensure}{\textbf{Output:}}
\begin{algorithm}[t]
    \caption{ARPE Algorithm}
    \label{alg:generate_random}
    \begin{algorithmic}[1]
    \REQUIRE Training Set $T$, number of selected reasoning patterns $r$, maximum number of exploration paths per pattern $n$, prompt $p$
    \ENSURE Dataset Generated by ARPE Strategy $D$
    
    \STATE $D'$ $\leftarrow \emptyset$;
    \STATE $patterns$ $\leftarrow \{ \}$; \texttt{// a dictionary, key: pattern, value: answer list}
    \STATE $patterns$ $\leftarrow$ groupByPattern($T$, $r$);
    \STATE \textbf{foreach} $pattern$ $\in$ $patterns$ \textbf{do}
        \STATE \quad $ans_{list}$ $\leftarrow$ $patterns$[$pattern$];
        \STATE \quad \textbf{foreach} $a$ $\in$ $ans_{list}$ \textbf{do}
            \STATE \quad \quad $paths$ $\leftarrow$ explorePaths($pattern$, $a$, $n$);
            \STATE \quad \quad \textbf{foreach} $path$ $\in$ $paths$ \textbf{do}
                \STATE \quad \quad \quad $sexpr$ $\leftarrow$ buildLogicalForm($path$);
                \STATE \quad \quad \quad $sparql$ $\leftarrow$ Convert($sexpr$);
                \STATE \quad \quad \quad \textbf{if} Execute($sparql$) is not $\emptyset$ \textbf{then}
                    \STATE \quad \quad \quad \quad $D'$.append(($path$, $sexpr$));

    \STATE $D$ $\leftarrow \emptyset$;
    \STATE \textbf{foreach} ($path$, $sexpr$) $\in$ $D'$ \textbf{do}
        \STATE \quad $q$ $\leftarrow$ LLMs($path$, $p$);
        \STATE \quad \textbf{if}  filter($q$) is True \textbf{then}
            \STATE \quad \quad $D$.append(($q$, $sexpr$));
    \STATE \textbf{return} $D$;
\end{algorithmic}
\end{algorithm}

Algorithm~\ref{alg:generate_random} outlines our ARPE strategy. Given a training set $T$, the function \verb|groupByPattern| clusters the original questions by their reasoning path patterns, sorts these clusters by the number of questions they contain, and returns the  top-$r$ clusters as $patterns$. For each pattern, we iterate over its answer list, treating each answer as an answer node and applying \verb|explorePaths| to perform reverse path exploration in the KG. This process yields up to $n$ paths per answer node. Next, \verb|buildLogicalForm| constructs the corresponding logical forms for these paths, which are then converted into SPARQL queries and executed to verify their correctness. Valid paths and their logical forms are stored in $D'$. Finally, we employ LLMs to generate questions for each path in $D'$ and use \verb|filter| to remove low-quality outputs, resulting in a dataset $D$ of \verb|(question, logical form)| pairs.

\begin{figure}[t]
	\centering	
	\includegraphics[width=\columnwidth]{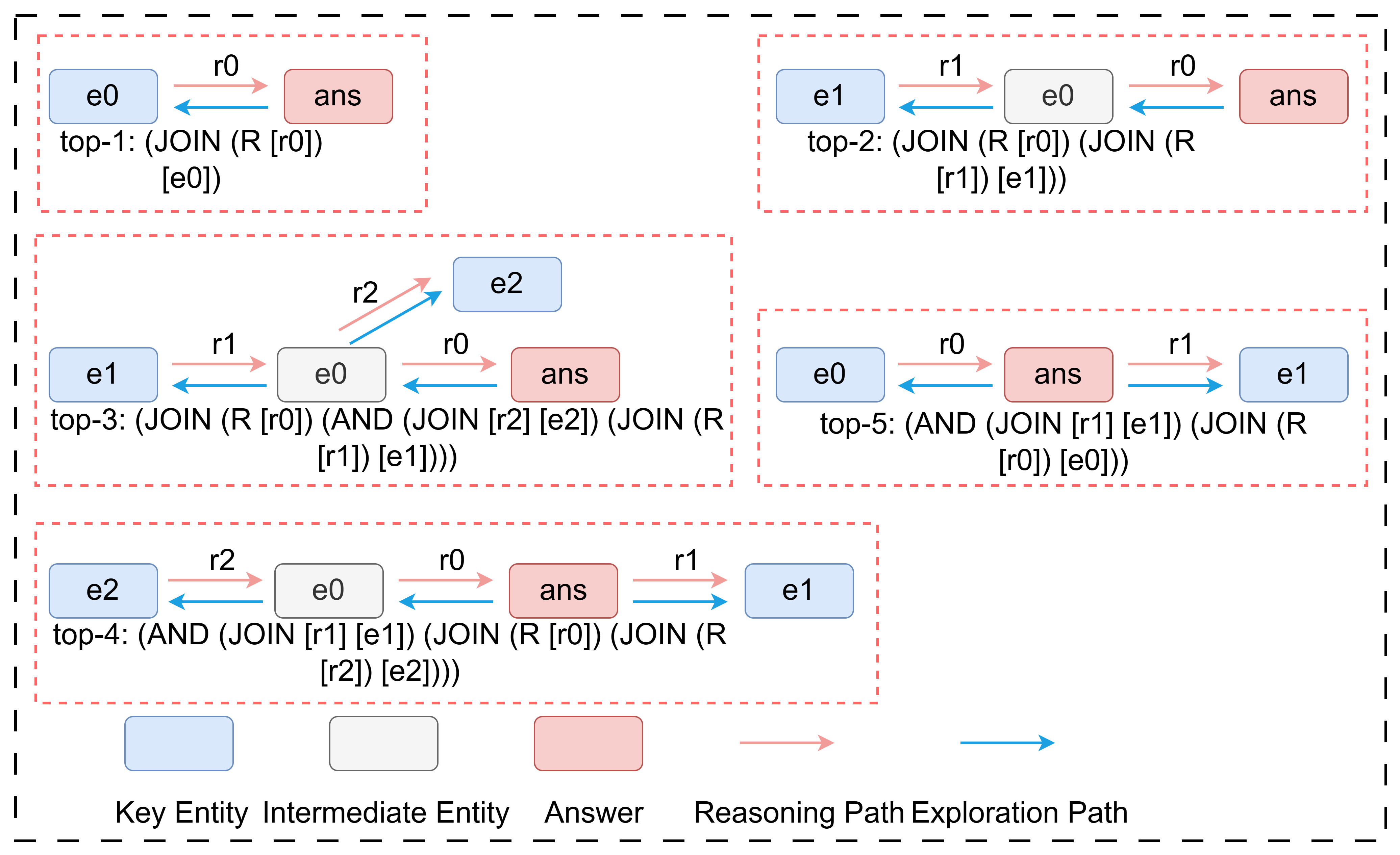} 
	\caption{\label{fig:WebQSP_random_path}Top-5 Reasoning Path Patterns in the WebQSP Training Set.}
\end{figure}

\begin{figure}[!h]
	\centering	
	\includegraphics[width=\columnwidth]{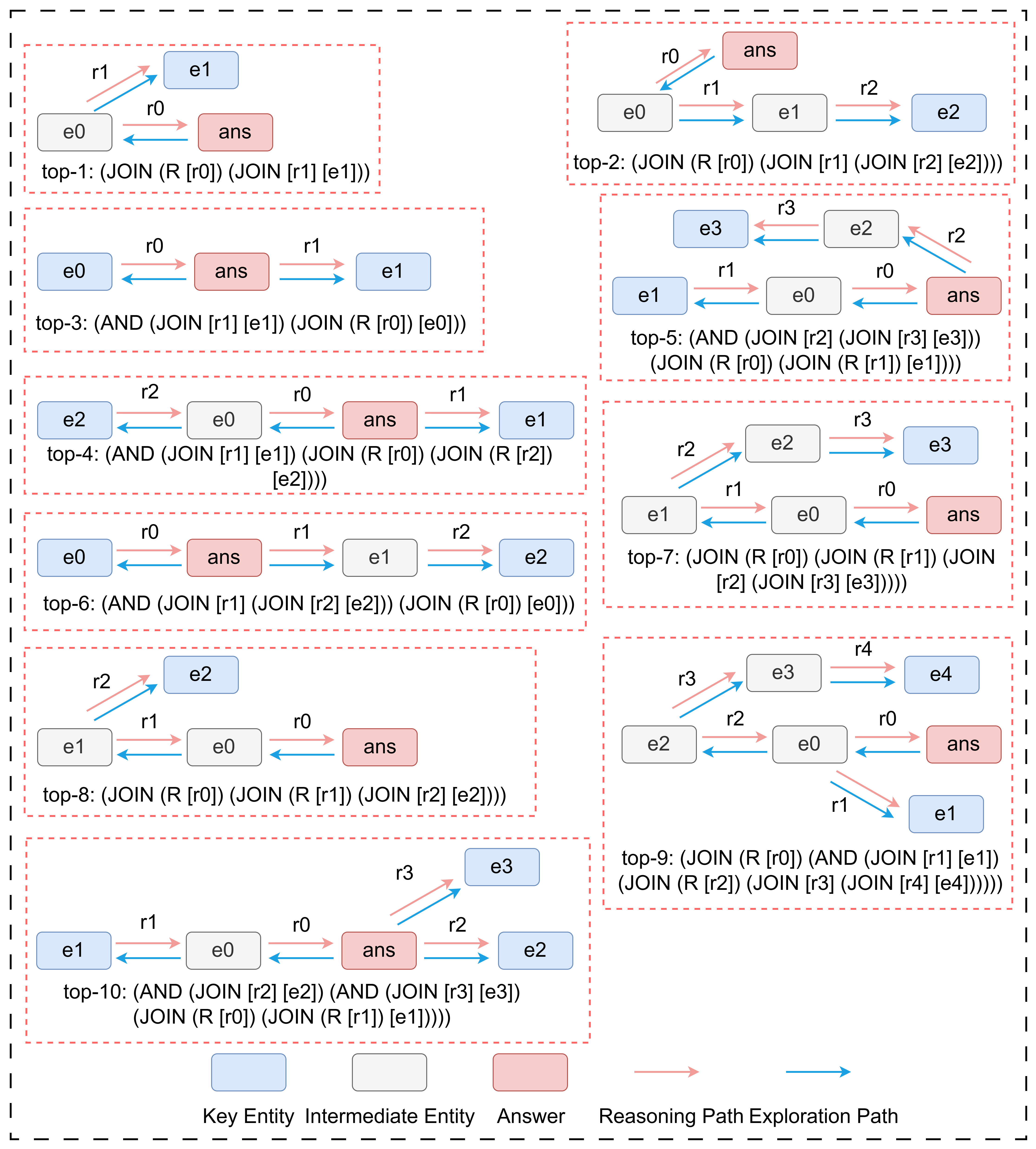} 
	\caption{\label{fig:CWQ_random_path}Top-10 Reasoning Path Patterns in the CWQ Training Set.}
\end{figure}

Figure~\ref{fig:WebQSP_random_path} and Figure~\ref{fig:CWQ_random_path} illustrate the Top-5 reasoning path patterns in the WebQSP training set and the Top-10 reasoning path patterns in CWQ, respectively. Specifically, ``Key Entity'' represents the key entities required for the reasoning path, which must be identified during path exploration. ``Intermediate Entity'' refers to entities within the reasoning path that do not require explicit identification. ``Answer'' serves as the starting point for reverse path exploration and corresponds to the final answer to the generated question. ``Reasoning Path'' denotes the relations between the head and tail entities in the KG. ``Exploration Path'' visualizes the exploration direction during path discovery, disregarding the actual relational direction.

\begin{figure}[t]
	\centering	
	\includegraphics[width=\columnwidth]{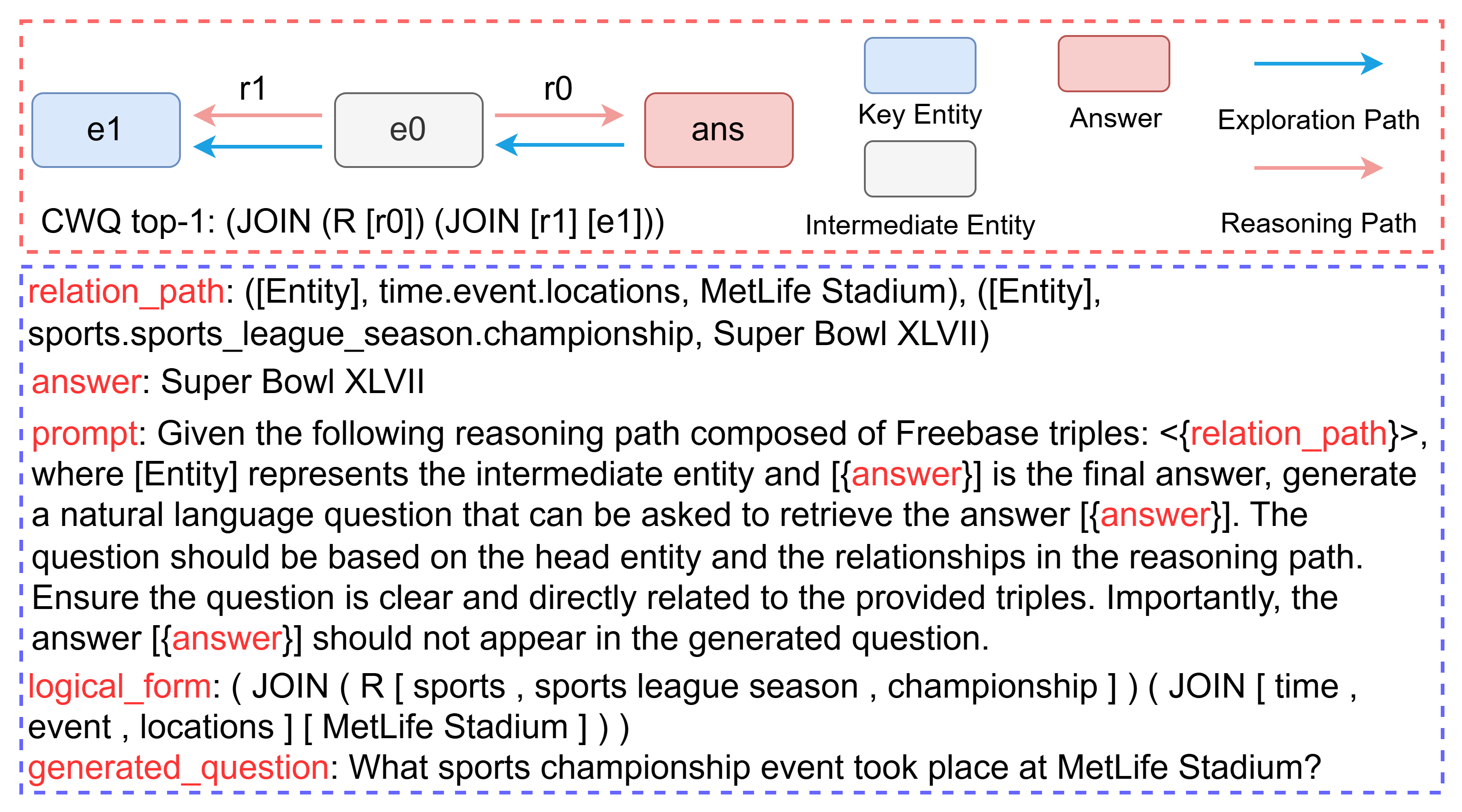} 
	\caption{\label{fig:arpe}An example of ARPE strategy.}
\end{figure}

As shown in Figure~\ref{fig:arpe}, a question is generated using the ARPE strategy based on the top-1 reasoning pattern from the CWQ dataset, with ``Super Bowl XLVII'' as its answer.

\subsection{Fine-Tuning on LLMs}

After generating three augmented datasets using our SPQG, SPQR, and ARPE strategies, we build the instruction fine-tuning dataset. We first convert SPARQL queries from the original dataset into S-expressions, replacing meaningless entity IDs with semantically rich labels (e.g., ``m.01y2hn6'' becomes ``School''). Next, we combine the instruction and questions as model inputs, with the processed S-expressions as the target logical forms. we use the parameter-efficient LoRA \cite{hu2022lora} method for fine-tuning, employing LLaMA2-7B and LLaMA2-13B \cite{touvron2023llama} as the base models for WebQSP and CWQ, respectively.

\subsection{Logical form generation}

Thanks to their strong text understanding and generation abilities, fine-tuned LLMs accurately follow instructions and produce well-structured logical forms. To assess our data augmentation strategies, we generate logical forms directly from the original questions without additional knowledge enhancements. When producing a single candidate, about 63.6\% of the outputs exactly match the ground truth. With beam search generating multiple candidates, accuracy rises to roughly 78.3\%. Moreover, after removing entities and relations (e.g., ``(JOIN (R []) [])''), the skeleton accuracy reaches approximately 92.1\%. These results indicate that fine-tuned LLMs possess strong semantic parsing capabilities and can generate high-quality logical forms.

\subsection{Unsupervised retrieval and interpretable execution}

Since fine-tuned LLMs generate accurate logical form skeletons, we follow \citet{luo2024chatkbqa} for unsupervised retrieval and replacement of entities and relations in candidate logical forms before converting the logical forms into SPARQL queries. This process further improves their accuracy.

Specifically, we first refine logical forms through entity retrieval and convert them into SPARQL queries to retrieve answers. For each question, we iterate over the generated logical forms. For each logical form $F$, we retrieve the top-$k$ most similar entities for each entity $e$ in $F$, using unsupervised methods like SimCSE \cite{gao2021simcse} or the FACC1 technique. The retrieved entities serve as candidate replacements for $e$. We then generate new logical forms by replacing the entities in $F$ with combinations of the retrieved similar entities, producing an updated list of logical forms, $F_{list}$. The final candidate list $C$ for the question consists of the refined logical forms $F_{list}$ generated for each $F$. Finally, each logical form in $C$ is converted into a SPARQL query and executed on the KG to obtain answers.

If no answers are retrieved from $C$, we apply a similar relation retrieval process to refine the logical forms in $C$, , generating a new candidate list $C'$. Finally, the logical forms in $C'$ are converted into SPARQL queries and executed on the KG to retrieve answers.

Overall, PGDA-KGQA utilizes open-source LLMs to construct augmented datasets using three strategies. With these data augmentation techniques, fine-tuned LLMs demonstrate improved semantic parsing. The generated logical forms are transformed into SPARQL queries and executed on the KG for answers. As SPARQL queries explicitly define the reasoning path, the inference process is transparent and interpretable.

\section{Experiments}

\subsection{Experimental setup}

\textbf{Datasets}. We conduct experiments on two standard KGQA datasets: WebQuestionsSP (WebQSP) \cite{yih2016value} and CWQ \cite{talmor2018web}, both based on Freebase \cite{bollacker2008freebase}. WebQSP contains 4,737 questions with corresponding SPARQL queries, while CWQ, with 34,689 questions, is derived by adding entities to WebQSP questions to create more complex queries. WebQSP queries typically require at most 2-hop reasoning, whereas CWQ involves more challenging 4-hop reasoning. We merge the training and validation sets and filter out questions whose SPARQL queries cannot be converted into S-expressions. After processing, the WebQSP training set includes 2,991 samples, while CWQ contains 30,416 samples.

\textbf{Baselines}. We compare PGDA-KGQA with a range of KGQA baseline models, including non-LLM-based methods such as EmbedKGQA \cite{saxena2020improving}, DECAF \cite{yu2022decaf}, as well as LLM-based methods such as ChatGPT and PanGu\cite{gu2023don}. Additional KGQA methods described in Section \ref{sec2} are also included.

\textbf{Evaluation Metrics}. We evaluate our models using F1 score, Hits@1, and Accuracy (Acc), following previous work \cite{shu2022tiara, luo2024chatkbqa}. Since some questions have multiple correct answers, the F1 score balances precision and recall to reflect overall answer coverage. Hits@1 indicates the proportion of test samples for which the top-ranked prediction is correct, and Acc is the percentage of questions with predictions that exactly match the ground truth.

\textbf{Hyperparameters and Environment}. All experiments use bfloat16 precision on NVIDIA 3090 GPUs (24GB). We fine-tune LLaMA2-7B on WebQSP and LLaMA2-13B on CWQ, as WebQSP has simpler questions and fewer samples. The WebQSP model is trained for 10 epochs on a single GPU with a learning rate of 2e-4, and a global batch size of 32. The CWQ model is trained for 10 epochs on two GPUs with a learning rate of 5e-4, a warmup ratio of 0.01, and a global batch size of 64. During generation, the beam size is set to 10 for WebQSP and 8 for CWQ.

\begin{table}[!t]
    \centering
    \adjustbox{width=\columnwidth,center}{
        \begin{tabular}{l|ccc|ccc}
        \toprule
        \multirow{2}{*}{\textbf{Model}} & \multicolumn{3}{c|}{\textbf{WebQSP}} & \multicolumn{3}{c}{\textbf{CWQ}} \\
        \cmidrule(lr){2-4} \cmidrule(lr){5-7}
        & F1 & Hits@1 & Acc & F1 & Hits@1 & Acc \\
        \midrule
        \multicolumn{7}{c}{non-LLM-based KGQA Methods} \\
        \midrule
        PullNet & --  &  68.1 & --  &  -- & 47.2 &  -- \\
        EmbedKGQA$^+$ & --  & 66.6  & --  & --  & 44.7  & --  \\
        QGG & 74.0  & 73.0  &  -- & 40.4  & 44.1  & --  \\
        CBR-KBQA & 72.8  & --  & 69.9  & 70.0  &  70.4 &  67.1 \\
        Subgraph Retrieval$^+$ & 64.1  & 69.5 & --  &  47.1 & 50.2  &  -- \\
        RnG-KBQA & 75.6  &  -- & 71.1  & --  & --  &  -- \\
        TIARA$^+$ & 78.9  &  75.2 & --  & --  & --  & --  \\
        GMT-KBQA & 76.6  & --  & 73.1  & 77.0  &  -- & 72.2  \\
        DECAF & 78.8  & 82.1  & --  & --  &  70.4 & --  \\
        BeamQA$^+$ & --  &  73.4 &  -- & --  &  -- &  -- \\
        FC-KBQA & 76.9  & --  & --  &  56.4 &  -- &  -- \\
        HGNet$^+$ & 76.6  & 76.9  & 70.7  &  68.5 &  68.9 &  57.8 \\
        \midrule
        \multicolumn{7}{c}{LLM-based KGQA Methods} \\
        \midrule
        ChatGPT$^+$ & 61.2  & --  & --  & 64.0  & --  & --  \\
        StructGPT$^+$ &  72.6 &  -- &  -- & --  &  -- &  -- \\
        Pangu & 79.6 &  -- & --  & --  &  -- & --  \\
        ToG &  -- & 82.6  & --  & --  &  69.5 & --  \\
        RoG & 70.8 & 85.7 &  -- & 56.2  &  62.6 & --  \\
        GSR$^+$ & 80.1  & 87.8  & --  & 64.4  & 67.5  &  -- \\
        ChatKBQA & 79.8  &  83.2 & 73.8  & 77.8  & 82.7  &  73.3 \\
        ChatKBQA$^+$ & 83.5  &  86.4 & 77.8  &  81.3 &  86.0 & 76.8  \\
        RGR-KBQA & 80.7  & 84.5  & 72.1  & 76.6  &  82.0 & 72.2  \\
        \midrule
        \multicolumn{7}{c}{Ours Methods} \\
        \midrule
        SPQG-KGQA & 83.1  & 86.0  & 77.7  & 79.7 & 84.0  & 75.6  \\
        SPQG-KGQA$^+$ & 85.6  &  88.3 &  80.7 &  82.9 & 87.0  & 78.9  \\
        \midrule
        SPQR-KGQA & 82.6  & 85.5  & 76.8  & 79.1 & 83.7  & 75.0  \\
        SPQR-KGQA$^+$ & \textbf{86.3}  &  \textbf{89.0} & \textbf{80.9} & 82.3 & 86.4  & 78.3  \\
        \midrule
        ARPE-KGQA &  81.8 & 85.2 & 75.1  & 79.2 & 83.7  & 75.2  \\
        ARPE-KGQA$^+$ &  85.0 & 87.9 & 79.3 & \textbf{83.1} & \textbf{87.1}  & \textbf{79.2}  \\
        \bottomrule
        \end{tabular}
    }
    \caption{KGQA result comparison of our methods with other baselines on WebQSP and CWQ datasets. $^+$ denotes using oracle entity linking annotations. The results of the models are mainly taken from their original paper. For our proposed methods, we display the results of the best setup on WebQSP and CWQ, respectively. The best results in each metric are in \textbf{bold}.}
    \label{tab:results}
\end{table}

\subsection{Results}

Table \ref{tab:results} compares the performance of PGDA-KGQA with baseline KGQA models on the WebQSP and CWQ datasets. The results show that all three data augmentation strategies yield performance improvements, consistently outperforming existing KGQA approaches on both datasets. Specifically, compared to previous best results, SPQG-KGQA improves F1, Hits@1, and Acc by 2.1\%, 0.5\%, and 2.9\% on WebQSP, and by 1.6\%, 1.0\%, and 2.1\% on CWQ. SPQR-KGQA achieves gains of 2.8\%, 1.2\%, and 3.1\% on WebQSP and 1.0\%, 0.4\%, and 1.5\% on CWQ. ARPE-KGQA yields improvements of 1.5\%, 0.1\%, and 1.5\% on WebQSP, and 1.8\%, 1.1\%, and 2.4\% on CWQ.

These results suggest that the three strategies contribute differently across dataset. On WebQSP, SPQR outperforms the others due to its ability to diversify question formulations, compensating for the limited sample size and enhancing generalization. SPQG improves the model's ability to align natural language with entities and relations by incorporating numerous single-hop questions. Although ARPE enhances the model's ability to handle questions following the Top-5 reasoning path patterns, its improvements are relatively modest, as cases involving rare entities or relations remain challenging.

Conversely, on CWQ, ARPE outperforms the other strategies, while SPQR offers the least benefit. The CWQ training set consists of 30,416 unique questions, derived from 3,042 original WebQSP questions, each expanded into about 10 variations. Consequently, CWQ inherently exhibits high linguistic diversity, which aligns with the principles of the SPQR strategy and limits its additional impact. In contrast, ARPE significantly improves performance by expanding the diversity of multi-hop reasoning paths, even though the quality of the generated questions is lower than that on WebQSP. SPQG improves the model by leveraging single-hop questions to help it decompose complex multi-hop queries into multiple relations.

\begin{table}[t]
    \centering
    \adjustbox{width=\columnwidth,center}{
        \begin{tabular}{cc|ccc|ccc}
        \toprule
        Dataset & Strategy & F1 & Hits@1 & Acc & F1$^+$ & Hits@1$^+$ & Acc$^+$ \\
        \midrule
        WebQSP & $\times$ & 81.0 & 84.3 & 74.9 & 84.2 & 87.0 & 78.8 \\
        CWQ & $\times$ & 78.2 & 82.5 & 74.2 & 81.4 & 85.5 & 77.5 \\
        \midrule
        \multicolumn{8}{c}{SPQG} \\
        \midrule
        \multirow{6}{*}{WebQSP} 
        & $k$=1$^-$ &  81.8 & 84.9 & 75.8 & 85.1 & 88.1 & 79.6 \\
        & $k$=1 &  81.8 & 84.9 & 76.4 & 84.6 & 87.2 & 79.7 \\
        & $k$=3 &  82.7 & 85.7 & 76.9 & 85.7 & 88.2 & 80.6 \\
        & $k$=5 &  \textbf{83.1} & \textbf{86.0} & \textbf{77.7} & 85.6 & \textbf{88.3} & \textbf{80.7} \\
        & $k$=7 &  82.2 & 85.1 & 76.6 & 85.3 & 87.9 & 80.2 \\
        & $k$=10 &  82.7 & 85.7 & 77.1 & \textbf{85.8} & \textbf{88.3} & 80.6 \\ \midrule
        \multirow{5}{*}{CWQ} 
        & $k$=1$^-$ & 78.7 & 82.9 & 74.8 & 81.8 & 85.7 & 78.1 \\
        & $k$=1 & 79.2 & 83.7 & 75.2 & 82.0 & 86.3 & 78.2 \\
        & $k$=2 & \textbf{79.7} & \textbf{84.0} & \textbf{75.6} & \textbf{82.9} & \textbf{87.0} & \textbf{78.9} \\
        & $k$=3 &  78.4 & 82.6 & 74.3 & 81.7 & 85.8 & 77.8 \\
        & $k$=10 &  78.2 & 82.6 & 74.1 & 81.5 & 85.7 & 77.4 \\
        \midrule
        \multicolumn{8}{c}{SPQR} \\
        \midrule
        \multirow{6}{*}{WebQSP} & $rw$=1 & \textbf{82.6} & \textbf{85.5} & \textbf{76.8} & \textbf{86.3} & \textbf{89.0} & \textbf{80.9} \\
        & $rw$=2 & 81.8 & 85.0 & 75.6 & 85.2 & 88.2 & 79.4 \\
        & $rw$=3 &  82.2 & 85.3 & 75.9 & 85.5 & 88.4 & 79.7 \\ 
        & $rw$=1$^*$ & 82.1 & 85.2 & 76.0 & 85.8 & 88.7 & 80.0 \\
        & $rw$=2$^*$ & 81.8 & 85.0 & 75.5 & 85.4 & 88.4 & 79.4  \\
        & $rw$=3$^*$ & 81.4 & 84.6 & 75.3 & 85.0 & 87.9 & 79.3  \\ \midrule
        \multirow{3}{*}{CWQ} & $rw$=1 & 78.6 & 82.7 & 74.7 & 81.9 & 85.9 & 78.1 \\
        & $rw$=3 & 78.0 & 81.9 & 74.3 & 80.7 & 84.4 & 77.2 \\ 
        & $rw$=1$^*$ & \textbf{79.1} & \textbf{83.7} & \textbf{75.0} & \textbf{82.3} & \textbf{86.4} & \textbf{78.3} \\
        \midrule
        \multicolumn{8}{c}{ARPE} \\
        \midrule
        \multirow{5}{*}{WebQSP} & \text{500} & 81.5 & 84.7 & \textbf{75.7} & 84.8 & 87.6 & 79.3 \\
        & \text{1000} & \textbf{81.8} & \textbf{85.2} & 75.1 & \textbf{85.0} & \textbf{87.9} & 79.3 \\
        & \text{2000} & 81.4 & 84.6 & 75.6 & 84.6 & 87.5 & 79.1 \\
        & \text{3000} & \textbf{81.8} & 85.0 & \textbf{75.7} & 84.8 & 87.7 & \textbf{79.4} \\
        & \text{All} & 81.7 & 84.9 & \textbf{75.7} & 84.8 & 87.6 & 
        \textbf{79.4}  \\ \midrule
        \multirow{5}{*}{CWQ} & \text{500} & 79.2 & 83.7 & 75.2 & \textbf{83.1} & \textbf{87.1} & \textbf{79.2} \\
        & \text{1000} & \textbf{79.5} & \textbf{84.0} & \textbf{75.3} & 82.6 & 86.8 & 78.5 \\
        & \text{2000} & 79.2 & 83.4 & 75.2 & 82.8 & 86.9 & 78.9 \\ 
        & \text{1000$^\#$} & 78.7 & 83.4 & 74.7 & 82.3 & 86.8 & 78.3 \\
        & \text{2000$^\#$} & 78.2 & 82.8 & 73.7 & 82.0 & 86.4 & 77.8 \\
        \bottomrule
        \end{tabular}
    }
    \caption{\label{tab:Ablation}Comparative performance of SPQG, SPQR, and ARPE on WebQSP and CWQ under varying augmentation settings. $^-$ denotes selecting only 1,000 samples. $^*$ denotes the use of samples belonging to the Top-5 reasoning path patterns. $^\#$ denotes the use of samples belonging to the Top-10 reasoning path patterns.}
\end{table}
\begin{table*}[!t]
    \centering
    \adjustbox{width=0.8\textwidth,center}{
	\begin{tabular}{ccccccccccc}
		\toprule
		Dataset & Top-1 & Top-2 & Top-3 & Top-4 &  Top-5 & Top-6 & Top-7 & Top-8 & Top-9 & Top-10  \\ \midrule
		WebQSP & 684 & 1500 & 2052 & 2843 & 4297 & -- & -- & -- & -- & -- \\
        CWQ & 3349 & 4253 & 4297 & 2843 & 4484 & 2836 & 4173 & 3280 & 3294 & 4479 \\ \midrule
        \multicolumn{11}{c}{Filter out low-quality questions} \\
        \midrule
        WebQSP & 573 & 1188 & 1741 & 2417 & 3859 & -- & -- & -- & -- & -- \\
		CWQ & 2225 & 3010 & 3859 & 2417 & 4018 & 2308 & 1800 & 2150 & 1675 & 1640 \\ \bottomrule
	\end{tabular}
    }
    \caption{\label{tab:random_paths_info}Question counts generated by ARPE Top-10 reasoning path patterns on WebQSP and CWQ.}
\end{table*}

\subsection{Ablation study}

We further conducted comprehensive ablation studies to answer three research questions. \textbf{RQ1:} How does the amount of augmented data, determined by different parameters in each strategy, affect model performance? (Sec \ref{sec: RQ1}) \textbf{RQ2:} Does the SPQR strategy enhance the model's generalization ability across different question formulations? (Sec \ref{sec: RQ2}) \textbf{RQ3:} Can the three strategies work together to further improve model performance? (Sec \ref{sec: RQ3})

\subsubsection{RQ1: Impact of augmented data quantity on model performance}
\label{sec: RQ1}

Table~\ref{tab:Ablation} compares the performance of SPQG, SPQR, and ARPE strategies across various augmentation settings on the WebQSP and CWQ datasets. For SPQG, the best improvements occur at $k$=5 on WebQSP and $k$=2 on CWQ, compared to the baseline. On WebQSP, performance levels off after $k$=3, suggesting that although single-hop questions improve generalization, their benefit eventually plateaus. This stable performance indicates that the generated questions are of high quality with minimal noise. In contrast, on CWQ, which requires multi-hop reasoning, a small number of single-hop questions help the model capture complex relations. However, excessive single-hop questions disrupt the original data distribution, reducing SPQG's effectiveness at higher $k$ values. 

For SPQR, the best results are achieved when only one paraphrase is generated per question ($rw$=1). As $rw$ increases, the quality of paraphrases declines, leading to diminishing performance gains. Moreover, testing with Top-5 reasoning path patterns shows a performance drop on WebQSP but an improvement on CWQ. Figure~\ref{fig:CWQ_random_path} explains that the Top-5 patterns in CWQ are relatively simple, resulting in high-quality paraphrases, while other complex patterns produce lower-quality paraphrases that limit the effectiveness of the SPQR strategy. For ARPE, using the Top-5 reasoning path patterns and restricting the number of generated questions to 500 per pattern results in significant gains on both datasets. Increasing this limit further does not benefit WebQSP and even worsens performance on CWQ due to the increased complexity of reasoning paths, which reduces question quality and introduces noise. Additionally, experiments with Top-10 patterns on CWQ confirm that low-quality questions from complex paths can negatively affect performance. Table~\ref{tab:random_paths_info} shows the number of samples generated by the ARPE strategy for both WebQSP and CWQ.

\subsubsection{RQ2: Effectiveness of SPQR in enhancing generalization}
\label{sec: RQ2}

To further evaluate the SPQR strategy's effectiveness in improving generalization across diverse question formulations, we applied SPQR to the WebQSP test set to generate a new test set and evaluated the model on it. Table~\ref{tab:Rewrite-Model-Ablation} compares the performance of the baseline model and the SPQR-enhanced model, indicating that SPQR improves nearly all metrics by at least 2\%, with accuracy increasing by approximately 3\%. These results confirm that introducing diverse paraphrases of equivalent semantics via SPQR effectively enhances the model's generalization capabilities.

\begin{table}[!ht]
    \centering
    \adjustbox{width=\columnwidth,center}{
        \begin{tabular}{c|ccc|ccc}
        \toprule
        Model & F1 & Hits@1 & Acc & F1$^+$ & Hits@1$^+$ & Acc$^+$ \\
        \midrule
        baseline  & 78.5 & 82.1 & 71.6 & 81.7 & 85.1 & 75.4 \\
        SPQR & \textbf{81.4} & \textbf{84.4} & \textbf{75.5} & \textbf{84.0} & \textbf{86.9} & \textbf{78.5} \\ 
        \bottomrule
        \end{tabular}
    }
    \caption{\label{tab:Rewrite-Model-Ablation}A comparison of the performance of the SPQR-enhanced model and the baseline on the WebQSP test set after applying the SPQR strategy.}
\end{table}

\subsubsection{RQ3: Can combining data augmentation strategies further improve model performance?}
\label{sec: RQ3}

\begin{table*}[!t]
    \centering
    \adjustbox{width=0.8\textwidth,center}{
        \begin{tabular}{cccc|ccc|ccc}
        \toprule
        Dataset & SPQG & SPQR & ARPE & F1 & Hits@1 & Acc & F1$^+$ & Hits@1$^+$ & Acc$^+$ \\
        \midrule
        WebQSP & $\times$ & $\times$ & $\times$ & 81.0 & 84.3 & 74.9 & 84.2 & 87.0 & 78.8 \\
        CWQ & $\times$ & $\times$ & $\times$ & 78.2 & 82.5 & 74.2 & 81.4 & 85.5 & 77.5 \\
        \midrule
        \multirow{9}{*}{WebQSP} & $k$=1 & $rw$=1 & $\times$ & 81.9 & 84.8 & 76.3 & 85.2 & 87.8 & 79.9 \\
         & $k$=1 & $rw$=3 & $\times$ & 82.2 & 85.4 & 75.8 & 85.6 & 88.6 & 79.7 \\
        & $k$=1 & $\times$ & \text{All} & 82.4 & 85.7 & 76.3 & 85.6 & 88.5 & 80.2 \\
        & $\times$ & $rw$=1 & \text{500} & 82.0 & 85.1 & 75.9 & 85.8 & 88.8 & 80.2 \\
        & $\times$ & $rw$=1 & \text{1000} & \textbf{83.5} & \textbf{86.8} & \textbf{77.3} & \textbf{86.3} & \textbf{89.2} & \textbf{80.8} \\
        & $\times$ & $rw$=1 & \text{2000} & 82.5 & 85.7 & 76.5 & 85.3 & 88.0 & 80.0 \\
        & $\times$ & $rw$=1 & \text{3000} & 82.3 & 85.4 & 76.1 & 85.3 & 88.2 & 79.6 \\
        & $\times$ & $rw$=1 & \text{All} & 82.4 & 85.7 & 76.1 & 85.4 & 88.2 & 79.9 \\
        & $k$=1$^-$ & $rw$=1 & \text{All} & 82.5 & 85.7 & 76.6 & 85.7 & 88.6 & 80.2 \\ \midrule
        \multirow{5}{*}{CWQ} & $k$=1$^-$ & $rw$=1$^*$ & $\times$ & \textbf{79.0} & 83.2 & \textbf{75.0} & 81.5 & 85.4 & 77.6 \\
        & $k$=1 & $rw$=1$^*$ & $\times$ & 78.9 & 83.3 & 74.9 & 81.7 & 85.7 & 77.9 \\
        & $k$=2 & $rw$=1$^*$ & $\times$ & 77.8 & 82.0 & 73.9 & 80.9 & 84.8 & 77.1 \\
        & $\times$ & \text{$rw$=1$^*$} & \text{500$^*$} & 78.4 & 83.0 & 74.3 & 81.3 & 85.9 & 77.1 \\
        & $k$=1 & $\times$ & \text{500$^*$} & \textbf{79.0} &\textbf{83.5} & 74.9 & \textbf{82.1} & \textbf{86.6} & \textbf{78.0} \\
        \bottomrule
        \end{tabular}
    }
    \caption{\label{tab:Fusion-Ablation}Performance of different strategy combinations on WebQSP and CWQ. $^-$ denotes selecting only 1,000 samples. $^*$ denotes the use of samples belonging to the Top-5 reasoning path patterns.}
\end{table*}

Table~\ref{tab:Fusion-Ablation} evaluates the impact of combining different data augmentation strategies. On the relatively simple WebQSP dataset, combining SPQG and SPQR yields minimal additional gains, whereas the combination of SPQR and ARPE further enhances performance when $rw$=1 and ARPE is capped at 1,000 samples per reasoning pattern. The most notable improvements occur when oracle entity linking annotations are excluded. In contrast, on the more complex CWQ dataset, the combination of different strategies fails to yield further improvements. As detailed in Sec~\ref{data_quality}, an analysis of the augmented samples suggests that, compared to WebQSP, the lower quality of LLM-generated questions for CWQ introduces noise, which accumulates when multiple strategies are applied together, ultimately reducing their effectiveness. While each strategy contributes to semantic understanding and generalization from distinct perspectives, their combined effect is constrained in noisy data environments, limiting further performance improvements.

\subsection{Analysis of augmented data quality}
\label{data_quality}

In the SPQR strategy, when reformulating based solely on the original question, increasing the value of $rw$ leads to high similarity between the generated reformulations, which can result in semantic deviation. For example, with the original question ``What was Malcolm X trying to accomplish'', the three generated reformulations are: ``What were Malcolm X's goals or objectives'', ``What did Malcolm X aim to achieve during his lifetime'', and ``What was the purpose behind Malcolm X's efforts and actions''. The first reformulation retains nearly the same meaning as the original, asking about Malcolm X's goals. The second reformulation introduces a time dimension (``lifetime''), which makes the question more specific but may not always align with the original intent. The third reformulation shifts focus to the motivation behind the actions, deviating from the core of the original question. This demonstrates that when $rw$ is large, relying solely on the original question and predefined prompts can introduce noise, reducing the model's performance. When $rw$ is small (e.g., 1 or 2), the generated reformulations remain more semantically consistent with the original, ensuring stability with less noise.

For complex multi-hop questions, the noise introduced during reformulation can sometimes outweigh the benefits. For example, in the CWQ dataset, the original question ``The actor who played in the film John Legend: Live from Philadelphia is engaged to whom?'' leads to three other reformulations such as ``Who is the actor who starred in the movie John Legend: Live from Philadelphia engaged to?'', ``To whom is the individual who portrayed a role in the film John Legend: Live from Philadelphia committed?'', and ``Which person is the performer from John Legend: Live from Philadelphia's film engaged to?''. In the first reformulation, ``starred in'' incorrectly limits the actor to being a lead. The second, using ``committed'' deviates from the core meaning of ``engaged''. The third reformulation, using ``performer'' blurs the actor's identity. Furthermore, the second reformulation's ``portrayed a role'' makes the expression unnecessarily verbose, reducing clarity.

\begin{figure}[!t]
	\centering	
	\includegraphics[width=\columnwidth]{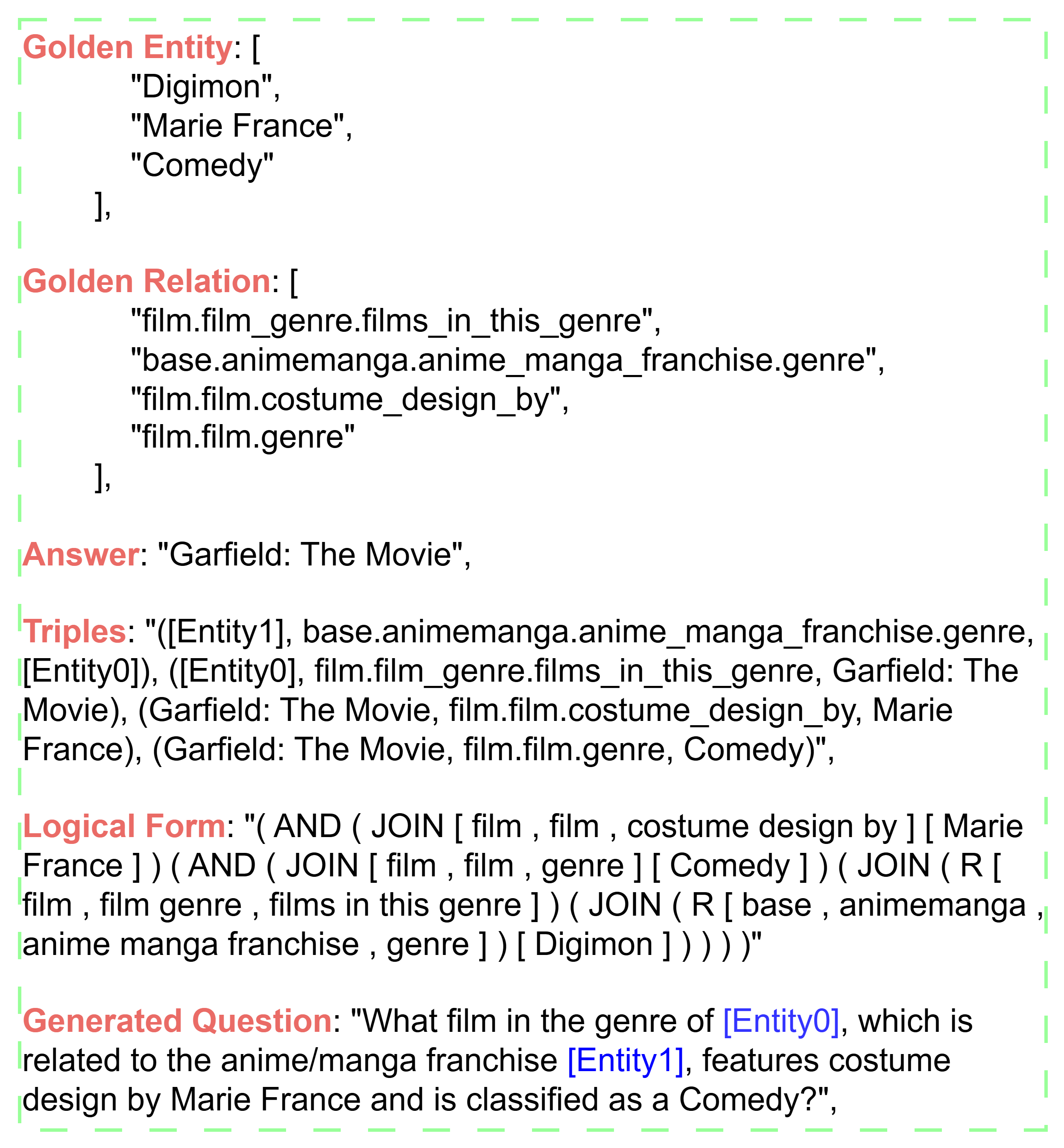} 
	\caption{\label{fig:low-quality}A low-quality question generated by ARPE strategy.}
\end{figure}

In the ARPE strategy, Table~\ref{tab:random_paths_info} presents the statistics of the questions generated using the ARPE strategy on the WebQSP and CWQ datasets. After filtering out low-quality questions (those containing placeholders like ``[Entity]'', KG relations, and questions with answers), we observe that the CWQ dataset has a significantly higher number of low-quality questions compared to WebQSP. As shown in the Figure~\ref{fig:low-quality}, one of the reasoning paths explored from the Top-10 reasoning patterns in the CWQ dataset, when used with LLMs to generate questions, resulted in questions containing intermediate entity placeholders like ``[Entity]''. This is unrealistic in practical scenarios, so these questions were filtered out to reduce the interference from low-quality data. This highlights that the current prompts and generation methods we are using are still insufficient when dealing with complex multi-hop questions.

\section{Conclusion}

In this paper, we introduce PGDA-KGQA, a prompt-guided augment-generate-retrieve framework that incorporates three data augmentation strategies with meticulously designed prompts: SPQG, SPQR, and ARPE. SPQG enhances the model's ability to parse complex questions by introducing a large number of single-hop questions, which improves the model's competence in handling sub-questions within multi-hop reasoning. SPQR improves generalization by rephrasing questions from the original dataset, increasing robustness to linguistic variations. ARPE specifically targets multi-hop reasoning by employing answer-guided reverse path exploration in the KG to generate complex multi-hop questions. Experimental results on WebQSP and CWQ demonstrate that PGDA-KGQA outperforms existing methods and achieves state-of-the-art performance in KGQA, validating the individual effectiveness of each strategy. These findings highlight that PGDA-KGQA not only enhances overall KGQA performance but also effectively addresses the challenges of multi-hop question answering through targeted data augmentation.

\section*{Limitations}

Although the SPQG, SPQR, and REPG strategies effectively enhance model performance by augmenting the training data, some noise remains in the generated questions, especially in complex multi-hop queries. These generated questions still lack naturalness and fluency. Future improvements can focus on the following areas: 
\begin{enumerate}
    \item Improving Prompt Design: Refining prompt templates to better utilize the generative capabilities of LLMs could produce more natural and high-quality questions. Additionally, introducing effective noise filtering mechanisms would enhance data quality and diversity, ensuring better alignment with real-world applications.
    \item Extension to Domain-Specific KGs: While the current framework focuses on general-domain knowledge graphs (e.g., Freebase), domain-specific knowledge graphs (e.g., in healthcare or law) also present valuable applications. Future work could adapt the proposed framework to these specific domains, addressing challenges such as complex entities, relationships, and high annotation costs.
\end{enumerate}

\bibliographystyle{elsarticle-num-names} 
\bibliography{reference}

\end{document}